\documentclass[runningheads]{llncs}
\usepackage{graphicx}
\usepackage{times}
\usepackage{latexsym}
\usepackage{graphicx}
\usepackage{url}
\usepackage{CJKutf8}
\usepackage{amsmath}
\usepackage{bm}
\usepackage{algorithm}
\usepackage{algorithmic}
\usepackage{url}
\usepackage[misc]{ifsym} 
\usepackage{bbding}

\begin{document}
\title{Automatic Translating between Ancient Chinese and Contemporary Chinese with Limited Aligned Corpora}
\titlerunning{Automatic Translating with Limited Aligned Corpora}
\author{
Zhiyuan Zhang\inst{1} \and
Wei Li\inst{1} \and
Qi Su\inst{2}\thanks{Qi Su is the corresponding author.}}
\authorrunning{Z. Zhang et al.}
\institute{
MOE Key Lab of Computational Linguistics, School of EECS,\\ Peking University, Beijing, China\\ \and
School of Foreign Languages, Peking University, Beijing, China \\ \email{\{zzy1210, liweitj47, sukia\}@pku.edu.cn}}
\maketitle              % typeset the header of the contribution
\begin{abstract}
The Chinese language has evolved a lot during the long-term development. Therefore, native speakers now have trouble in reading sentences written in ancient Chinese. In this paper, we propose to build an end-to-end neural model to automatically translate between ancient and contemporary Chinese. However, the existing ancient-contemporary Chinese parallel corpora are not aligned at the sentence level and sentence-aligned corpora are limited, which makes it difficult to train the model. To build the sentence level parallel training data for the model, we propose an unsupervised algorithm that constructs sentence-aligned ancient-contemporary pairs by using the fact that the aligned sentence pair shares many of the tokens. Based on the aligned corpus, we propose an end-to-end neural model with copying mechanism and local attention to translate between ancient and contemporary Chinese. Experiments show that the proposed unsupervised algorithm achieves 99.4\% F1 score for sentence alignment, and the translation model achieves 26.95 BLEU from ancient to contemporary, and 36.34 BLEU from contemporary to ancient.
\end{abstract}

\section{Introduction}
% 加引用 attention， LSTM双向， dropout，NMT，copy相关的
% 先说数据缺乏的问题，再说需要构建数据， 把两个工作联系起来，放例子解释对齐和翻译用到的特性,IBM Model 句子对齐引用
The ancient Chinese was used for thousands of years. There is a huge amount of books and articles written in ancient Chinese. However, both the form and grammar of the ancient Chinese have been changed. Chinese historians and littérateurs have made great efforts in translating such literature into contemporary Chinese, a big part of which is publicly available on the Internet. However, there is still a big gap between these literatures and parallel corpora, because most of the corpora are coarsely passage-aligned, the orders of sentences are different. To train an automatic translating model, we need to build a sentence-aligned corpus first.

Translation alignment is an important pre-step for machine translation. Most of the previous work focuses on how to apply supervised algorithms on this task using features extracted from texts. The Gale Church algorithm~\cite{gale,haruno_yamazaki_1997} uses statistical or dictionary information to build alignment corpus. Strands~\cite{Resnik1998Parallel,conf/acl/Resnik99} extracts parallel corpus from the Internet. A logarithmic linear model~\cite{japanese-chinese} is also used for Chinese-Japanese clause alignment. Besides, features such as sentence lengths, matching patterns, Chinese character co-occurrence in Japanese and Chinese are also taken into consideration. The observation that Chinese character co-occurrence also exists in ancient-contemporary Chinese is also used to ancient-contemporary Chinese translation alignment~\cite{Lin2007,an_con}.

The method above works well, however, these supervised algorithms require a large parallel corpus to train, which is not available in our circumstance. These previous algorithms did not make good use of the characteristics of ancient-contemporary Chinese pair. To overcome these shortcomings, we design an unsupervised algorithm for sentence alignment based on the observation that differently from a bilingual corpus, ancient-contemporary sentence pairs share many common characters in order.
We evaluate our alignment algorithm on an aligned parallel corpus with small size. The experimental results show that our simple algorithm works very well (F1 score 99.4), which is even better than the supervised algorithms. 

Deep learning has achieved great success in tasks like machine translation. The sequence to sequence (seq-to-seq) model~\cite{seq2seq} is proposed to generate good translation results on machine translation. The attention mechanism~\cite{bahdanau2014neural} is proposed to allow the decoder to extract phrase alignment information from the hidden states of the encoder.
Most of the existing NMT systems are based on the seq-to-seq model~\cite{Kalchbrenner,ChoEA2014,seq2seq} and the attention mechanism.
Some of them have variant architectures to capture more information from the inputs~\cite{lattice,Copy}, and some improve the attention mechanism~\cite{stanfordattention,interactive,supervisedattention,mapattention,fengattention,doublyattention}, which also enhanced the performance of the NMT model. Inspired by the work of pointer generator~\cite{pointer}, we use a copying mechanism named pointer-generator model to deal with the task of ancient-contemporary translation task because ancient and contemporary Chinese share common characters and some same proper nouns.

Experimental results show that a copying mechanism can improve the performance of seq-to-seq model remarkably on this task. We show some experimental results in the experiment section. Other mechanisms are also implied to improve the performance of machine translation~\cite{DBLP:Lin:2018,DBLP:Ma:2018}.

Our contributions lie in the following two aspects:
\begin{itemize}
\item We propose a simple yet effective unsupervised algorithm to build the sentence-aligned parallel corpora, which make it possible to train an automatic translating model between ancient Chinese and contemporary Chinese even with limited sentence-aligned corpora.
\item We propose to apply the sequence to sequence model with copying mechanism and local attention to deal with the translating task. Experimental results show that our method can achieve the BLEU score of 26.41 (ancient to contemporary) and 35.66 (contemporary to ancient).
\end{itemize}

\section{Proposed Method}

\subsection{Unsupervised Algorithm for Sentence Alignment}

We review the definition of sentence alignment first. 

Given a pair of aligned passages, the source language sentences $S$ and target language sentences $T$, which are defined as
\begin{align}
S:=\{s_1,s_2,\cdots ,s_n\},\quad T:=\{t_1,t_2,\cdots ,t_m\}
\end{align}

The objective of sentence alignment is to extract a set of matching sentence pairs out of the two passages. Each matching pair consists of several sentence pairs like $(s_i, t_j)$, which implies that $s_i$ and $t_j$ form a parallel pair.

For instance, if $S=\{s_1, s_2, s_3, s_4, s_5\}, T=\{t_1, t_2, t_3, t_4, t_5\}$ and the alignment result is $\{(s_1, t_1), (s_3, t_2), (s_3, t_3), (s_4, t_4), (s_5, t_4)\}$. It means source sentence $s_1$ is translated to target sentence $t_1$, $s_3$ is translated to $t_2$ and $t_3$, $s_4$ and $s_5$ are translated to $t_4$ while $s_2$ is not translated to any sentence in target sentences. Here we may assume that source sentences are translated to target sentences in order. For instance $(s_1, t_2)$ and $(s_2, t_1)$ can not exist at the same time in the alignment result.

Translating ancient Chinese into contemporary Chinese has the characteristic that every word of ancient Chinese tends to be translated into contemporary Chinese in order, which usually includes the same original character. Therefore, correct aligned pairs usually have the maximum sum of lengths of the longest common subsequence (LCS) for each matching pair. 

Let $\textbf{lcs}(s[i_1, i_2], t[j_1,j_2])$ be the length of the \textbf{longest common subsequence} of a matching pair of  aligned sentences consisting of source language sentences $s[i_1,i_2]$ and target language sentences $t[j_1,j_2]$, which are defined as
\begin{equation}
s[i_1,i_2]=\left\{
\begin{aligned}
s_{i_1}s_{i_1+1}\cdots s_{i_2} (i_1\le i_2) \\
[empty\_str]\ (i_1>i_2)
\end{aligned}
\right.
,\quad
t[j_1,j_2]=\left\{
\begin{aligned}
t_{j_1}t_{j_1+1}\cdots t_{j_2} (j_1\le j_2) \\
[empty\_str]\ (j_1>j_2)
\end{aligned}
\right.
\end{equation}
where $[empty\_str]$ denotes an empty string. The longest common subsequence of an empty string and any string is $0$.

%%%%%%%%%%%%%%%%%%%%%%%%%%%%%%%%%%%%%%%%%%%%%%%%%%%
We use the dynamic programming algorithm to find the maximum score and its corresponding alignment result. 
Let $f(i, j)$ be the maximum score that can be achieved with partly aligned sentence pairs until $s_i$, $t_j$. To reduce the calculation cost, we only consider cases where one sentence is matched with no more than 5 sentences:
\begin{align}
f(i, j)= \mathop{max} \limits_{\{i-i', j-j'\}=\{1, M\}, 0\le M \le 5}\{f(i', j') + \textbf{lcs}(s[i'+1, i], t[j'+1, j])\}
\end{align}
%%%%%%%%%%%%%%%%%%%%%%%%%%%%%%%%%%%%%%%%%%%%%%%%%%%

%%%%%%%%%%%%%%%%%%%%%%%%%%%%%%%%%%%%%%%%%%%%%%%%%
%We use the dynamic programming algorithm to find the maximum score and its corresponding alignment result. 
%Let $f(i, j)$ be the maximum score that can be achieved with partly aligned sentence pairs until $s_i$, $t_j$ respectively:
%\begin{align}
%f(i, j)= \mathop{max} \limits_{0<i'\le i, 0<j'\le j}\{f(i', j') + \textbf{lcs}(s[i'+1, i], t[j'+1, j])\}
%\end{align}

%We can preprocess all $\textbf{lcs}(s[i'+1, i], t[j'+1, j])$ scores and store them for using in algortihm, which has a time complexity of $O\big((mn)^2\big)$. Here we enumerate every $i',j'$ that the maximum value of $f(i, j)$ is achieved when $s_{i'+1}\cdots s_{i}$ and $t_{j'+1}\cdots t_{j}$ are aligned. We need to calculate $f(i,j)$ for $O(nm)$ times. For every $f(i, j)$, we need to enumerate $i',j'$ for $O(nm)$ times. The time complexity of this dynamic programming algorithm is $O\big((nm)^2\big)$, which is too costly when $n$ and $m$ are large.

%To reduce the calculation cost, we only consider cases where one sentence is matched with no more than 5 sentences, namely $\{i-i', j-j'\}=\{1, M\}, 0\le M \le 5$ ($M\ge 0$ holds if and only if one sentence is not matched with none of the sentences). Namely
%\begin{align}
%f(i, j)= \mathop{max} \limits_{\{i-i', j-j'\}=\{1, M\}, 0\le M \le 5}\{f(i', j') + \textbf{lcs}(s[i'+1, i], t[j'+1, j])\}
%\end{align}
%%%%%%%%%%%%%%%%%%%%%%%%%%%%%%%%%%%%%%%%%%%%%%%

%%%%%%%%%%%%%%%%%%%%%%%%%%%%%%%%%%%%%%%%%%%%%%%%%%%%%%%%%%%%%
\begin{algorithm}[t]
\caption{Dynamic Programming Algorithm to Find the Alignment Result.}
\label{algorithm1}
\begin{algorithmic}
\STATE Preprocess \textbf{lcs} scores and store them in memory. Initialize $f[i, j] \gets 0$.\\
\STATE \textbf{for} $i\in [1,n]$\\
\STATE \quad \textbf{for} $j\in [1,m]$\\
\STATE \quad \quad \textbf{for} $i',j'\in \{(i', j'):\{i-i', j-j'\}=\{1, M\}, 0\le M \le 5, i'>0, j'>0 \}$\\
\STATE \quad \quad \quad Get $now\_lcs\gets\textbf{lcs}(s[i'+1, i], t[j'+1, j])$ from memory.\\
\STATE \quad \quad \quad \textbf{if} $f[i',j']+now\_lcs > f[i, j]$ \\
\STATE \quad \quad \quad \quad $f[i,j]=f[i',j']+now\_lcs$ \\
\STATE \quad \quad \quad \quad Update the corresponding alignment result.\\
\STATE \quad \quad \quad \textbf{endif}
\STATE \quad \quad \textbf{endfor}
\STATE \quad \textbf{endfor} 
\STATE \textbf{endfor} 
\end{algorithmic}
\end{algorithm}
%%%%%%%%%%%%%%%%%%%%%%%%%%%%%%%%%%%%%%%%%%%%%%%%%%%%%%%%%%%%%%%

The condition $\{i-i', j-j'\}=\{1, M\}, 0\le M \le 5$ ensures that one sentence is matched with no more than 5 sentences and  $M = 0$ holds if and only if one sentence is not matched with any of the sentences. We can preprocess all $\textbf{lcs}(s[i'+1, i], t[j'+1, j]) (\{i-i', j-j'\}=\{1, M\}, 0\le M \le 5)$ scores and store them for using in algorithm, which has a time complexity of $O(mn)$. The pseudo code is shown in Algorithm~\ref{algorithm1}. Then for every $f(i, j)$, we only need to enumerate $i',j'$ for $O(1)$ times and the time complexity of proposal dynamic programming algorithm is $O(mn)$. 

When the size of corpus grows, this algorithm will be time-consuming, which means our algorithm is more suitable for passage-aligned corpora instead of a huge text. In reality, we find that when translating a text written in ancient Chinese into contemporary Chinese. Translators tend not to change the structure of the text, namely, a book written in ancient Chinese and its contemporary Chinese version are a passage-aligned corpus naturally. Therefore, we can get abundant passage-aligned corpora form Internet.

\subsection{Neural Machine Translation Model}
Sequence-to-sequence model was first proposed to solve machine translation problem. The model consists of two parts, an encoder and a decoder. The encoder is bound to take in the source sequence and compress the sequence into hidden states. The decoder is used to produce a sequence of target tokens based on the information embodied in the hidden states given by the encoder. Both encoder and decoder are implemented with Recurrent Neural Networks (RNN). 

To deal with the ancient-contemporary translating task, we use the encoder to convert the variable-length character sequence into $h_t$, the hidden representations of position $t$, with an Bidirectional RNN:
\begin{align}
\stackrel{\rightarrow}{h_t} = f(x_t, \stackrel{\rightarrow}{h_{t-1}}),\quad
\stackrel{\leftarrow}{h_t} = f(x_t, \stackrel{\leftarrow}{h_{t+1}}), \quad
{h_t} = [\stackrel{\rightarrow}{h_t}; \stackrel{\leftarrow}{h_t}]
\end{align}
Where $f$ is a function of RNN family, $x_t$ is the input at time step $t$. The decoder is another RNN, which generates a variable-length sequence token by token, through a conditional language model,
\begin{align}
s_t = f(c_t, s_{t-1}, Ey_{t-1}),\quad
c_t = g(\bm{h}, s_{t-1})
\end{align}
Where $E$ is the embedding matrix of target tokens, $y_{t-1}$ is the last predicted token. We also implement a beam search mechanism, a heuristic search algorithm that expands the most promising node in a limited set, for generating a better target sentence. 

In the decoder, suppose $l$ is the length of the source sentence, then the context vector $c_t$ is calculated based on the hidden states $s_t$ of the decoder at time step $t$ and all the hidden states $\bm{h}=\{h_i\}_{i=1}^l$ in the encoder, which is also known as the attention mechanism,
\begin{align}
\beta_{t,i} = v_a^\text{T}\tanh(W_as_t + U_ah_i + b_a),\quad
a_{t,i}  = \frac{\exp(\beta_{t,i})}{\sum\limits_{j=1}^{l}\exp(\beta_{t,j})}
\end{align}

We adopt the global attention mechanism in the baseline seq-to-seq model,
\begin{align}
c_t = \sum\limits_{i=1}^{l}a_{t,i}h_i
\end{align}
where $a_{t,i}$ is the attention probability of the word in  position $i$ of the source sentence in the decoder time step $t$ and $h_i$ is the hidden state of position $i$ of the source sentence.

Because in most cases ancient and contemporary Chinese have similar word order, instead of the normal global attention in the baseline seq-to-seq model, we apply \textit{local attention}~\cite{stanfordattention,Tjandra2017Local} in our proposal. When calculating the context vector $c_t$, we calculate a pivot position in the hidden states $\bm{h}$ of the encoder, and calculate the attention probability in the window around the pivot instead of the whole sentence, 
\begin{align}
c_t &= \sum\limits_{i=1}^{l}p_{t,i}a_{t,i}h_i
\end{align}
where $p_{t,i}$ is the mask score based on the pivot position in the hidden states $\bm{h}$ of the encoder.

Machine translation model treats ancient and contemporary Chinese as two languages, however, in this task, contemporary and ancient Chines share many common characters. Therefore, we treat ancient and contemporary Chinese as one language and share the character embedding between the source language and target language.

\subsection{Copying Mechanism}
As is stated above, ancient and contemporary Chinese share many common characters and most of the name entities use the same representation. Copying mechanism~\cite{gu2016incorporating} is very suitable in this situation, where the source and target sequence share some of the words. We apply pointer-generator framework in our model, which follows the same intuition as the copying mechanism. $p_t(w)$, the output probability of token $w$ is calculated as,
\begin{align}
p_t(w) = p^G_tp^V_t(w) + (1-p^G_t)\sum\limits_{\text{src}_i=w}a_{t,i},\quad
p^G_t = \sigma(W_gs_t + U_gc_t + b_g)
\end{align}
where $p^G_t$ is dynamically calculated based on the hidden state $s_{t}$ and the context vector $c_t$, $P^V_t(w)$ is the probability for token $w$ in traditional seq-to-seq model. $a_{t,i}$ is the attention probability at the time step $t$ in decoder and position $i$ in source sentence which satisfies $\text{src}_i=w$, namely the token in position $i$ in source sentence is $w$. $\sigma$ is the sigmoid function.

The encoder and decoder networks are trained jointly to maximize the conditional probability of the target sequence. We use cross-entropy as the loss function. We use characters instead of words because characters have independent meaning in ancient Chinese and the number of characters is much lower than the number of words, which makes the data less sparse and greatly reduces the number of OOV. We also can implement the pointer-generator mechanism in summarization on our AT-seq-to-seq model to copy some proper nouns directly from the source language.

%%%%%%%%%%%%%%%%%%%%%%%%%%%%%%%%%%%%%%%%%%%%%%%%%%%%
\begin{table}[t]
\caption{Vocabulary statistics. We include all the characters in the training set in the vocabulary.}
%\footnotesize
\centering
\begin{tabular}{|c|c|c|}
\hline
\label{tab:vocab}
\bf Language & \bf Vocabulary & \bf OOV Rate \\ \hline
Ancient & 5,870 & 1.37\%\\
Contemporary & 4,993 & 1.03\% \\ \hline
\end{tabular}
\end{table}
%%%%%%%%%%%%%%%%%%%%%%%%%%%%%%%%%%%%%%%%%%%%%%%%%%%%

\section{Experiments}

\subsection{Datasets}

\subsubsection*{Sentence Alignment}

We crawl passages and their corresponding contemporary Chinese version from the Internet. To guarantee the quality of the contemporary Chinese translation, we choose the corpus from two genres, classical articles and Chinese historical documents. 
After proofreading a sample of these passages, we think the quality of the \textbf{passage-aligned} corpus is satisfactory.
To evaluate the algorithm, we crawl a relatively small \textbf{sentence-aligned} corpus consisting of 90 aligned passages with 4,544 aligned sentence pairs. We proofread them and correct some mistakes to guarantee the correctness of this corpus.

\subsubsection*{Translating}

We conduct experiments on the data set built by our proposed unsupervised algorithm. The data set consists of 57,391 ancient-contemporary Chinese sentence pairs in total. We split sentence pairs randomly into train/dev/test dataset with sizes of 53,140/2,125/2,126 respectively. The vocabulary statistics information is in Table~\ref{tab:vocab}.

\subsection{Experimental Settings}

\subsubsection*{Sentence Alignment}

We implement a log-linear model on contemporary-ancient Chinese sentence alignment as a baseline. Following the previous work, we implement this model with a combination of three features, sentence lengths, matching patterns and Chinese character co-occurrence~\cite{japanese-chinese,Lin2007,an_con}. We split the data into training set (2,999) and test set (1,545) to train the log-linear model. Our unsupervised method does not need training data.  Both these two methods are evaluated on the test set. 

\subsubsection*{Translating}

We conduct experiments on both translating directions and use BLEU score to evaluate our model. We implement a one-layer Bidirectional LSTM with a 256-dim embedding size, 256-dim hidden size as encoder and a one-layer LSTM with attention mechanism as decoder. We also implement local attention or pointer generator mechanism on the decoder in our proposed model. We adopt Adam optimizer to optimize our loss functions with a learning rate of 0.0001. The training batch size of our model is 64 and the generating batch size is 32. The source language and target language share the vocabulary. 
%%%%%%%%%%%%%%%%%%%%%%%%%%%%%%%%%%%%%%%%%%%%%%%%%%%%%%%%%%%%%%%%%%%%%

\begin{table}[t]
\caption{Evaluation of logarithmic linear models and our method.}
\label{tab:alignment result}
%%\footnotesize
\begin{center}
\begin{tabular}{|l|c|c|c|}
\hline
\bf Features& \bf Precision & \bf Recall & \bf F1-score \\
\hline
   Length &   0.821 &0.855 & 0.837\\
   Pattern  &  0.331 & 0.320 & 0.325\\
   Length and Pattern  &  0.924& 0.912 & 0.918\\
   Co-occurrence  &  0.982 & 0.984 & 0.983\\
   Length and Co-occurrence  &  0.987& 0.989 & 0.988\\
   Pattern and Co-occurrence  &  0.984& 0.980 & 0.982\\
   All features & 0.992 & 0.991 & 0.992\\
   \hline
    our method & \textbf{0.994} & \textbf{0.994} & \textbf{0.994} \\
    \hline
\end{tabular}
\end{center}
\end{table}
\begin{table}[t]
%\footnotesize
\caption{Evaluation result (BLEU) of translating between \textbf{An} (ancient) and \textbf{Con} (contemporary) Chinese in test dataset}
\begin{center}
\begin{tabular}{|l|c|c|}
\hline
\bf Method & \bf An-Con & \bf Con-An \\
\hline
Seq-to-Seq & 23.10 & 31.20 \\
+ copying & 26.41 & 35.66 \\
+ Local Attention & \textbf{26.95} & \textbf{36.34} \\
\hline
\end{tabular}
\end{center}
\label{tab:result}
\end{table}
%%%%%%%%%%%%%%%%%%%%%%%%%%%%%%%%%%%%%%%%%%%%%%%%%%%%%

\subsection{Experimental Results}

\subsubsection*{Sentence Alignment}
Our unsupervised algorithm gets an F1-score of 99.4\%, which is better than the supervised baseline with all features, 99.2\% (shown in Table \ref{tab:alignment result}). Our proposal algorithm does not use the features of sentence lengths and matching pattern. If we only adopt the feature of Co-occurrence, the performance of our proposal is 1.1\% higher on F1-score than that of the supervised baseline (98.3\%). To conclude, our proposal algorithm uses fewer features and use no training data but perform better than the supervised baseline.

\subsubsection*{Translating}
Experimental results show our model works well on this task (Table \ref{tab:result}). Compared with the basic seq-to-seq model, copying mechanism gives a large improvement, because the source sentences and target sentence share some common representations, we will also give an example in the case study section. Local attention gives a small improvement over the traditional global attention, this can be attributed to shrinking the attention range, because most of the time, the mapping between ancient and contemporary Chinese is clear. A more sophisticated attention mechanism which makes full use of the characteristics of ancient and contemporary Chinese may further improve the performance.

\section{Discussion}

\subsection{Sentence Alignment}

We find that the small fraction of data (0.6\%) that our method makes mistakes are mainly because of the change of punctuation. For example, in ancient Chinese, there is a comma ``,'' after ``\begin{CJK}{UTF8}{gbsn}异哉，\end{CJK}'' (How strange!), while in contemporary Chinese, ``\begin{CJK}{UTF8}{gbsn}怪啊！\end{CJK}'' (How strange!), an exclamation mark ``!'' is used, which makes the sentence to be an independent sentence. Since the sentence is short and there is no common character, our method fails to align the sentences correctly. However, such problems also exist in supervised models.

\subsection{Necessity of Building a Large Sentence-aligned Corpus}

From Table~\ref{tab: small train result}, we can see that the results are very sensitive to the scale of the training data size. Therefore, our unsupervised method of building a large sentence-aligned corpus is necessary. If we do not build a large sentence-aligned corpus by our sentence alignment algorithm, we will only have limited sentence-aligned corpora and the performance of translating will be worse.

%%%%%%%%%%%%%%%%%%%%%%%%%%%%%%%%%%%%%%%%%%%%%%%%%%%%%%
\begin{table}[t]
%\footnotesize
\caption{Evaluation result (BLEU) of translating between \textbf{An} (ancient) and \textbf{Con} (contemporary) Chinese with different number of training samples.}
\begin{center}
\begin{tabular}{|l|c|c|c|c|}
\hline
\bf Language & 5,000 & 10,000 & 20,000 & 53,140 \\
\hline
An-Con & 3.00 &  9.69 & 16.31 & 26.95\\
Con-An & 2.40 & 10.14 & 18.47 & 36.34\\
\hline
\end{tabular}
\end{center}
\label{tab: small train result}
\end{table}
\begin{table}[t]
%\footnotesize
\caption{Example of translating from \textbf{An} (ancient) to \textbf{Con} (contemporary) Chinese.}
\begin{center}
\begin{tabular}{|l|p{6cm}|}
\hline
{\bf Source} & \begin{CJK}{UTF8}{gbsn}六月辛卯，中山王焉薨。\end{CJK} \\ (Translation in English) &
(On Xinmao Day of the sixth lunar month, Yan, King of Zhongshan, passed away.) \\ \hline
{\bf Target} & \begin{CJK}{UTF8}{gbsn} 六月十二日，中山王刘焉逝世。\end{CJK} \\ (Translation in English) &
(On twelfth of the sixth lunar month, Liu Yan, King of Zhongshan, passed away.) \\ \hline
{\bf Seq2Seq} & \begin{CJK}{UTF8}{gbsn}六月十六日，中山王刘裕去世。\end{CJK} \\ (Translation in English) &
(On sixteenth of the sixth lunar month, Liu Yu, King of Zhongshan, died.) \\ \hline
{\bf Our model} & \begin{CJK}{UTF8}{gbsn}六月二十二日，中山王刘焉逝世。\end{CJK} \\ (Translation in English) &
(On twenty-second of the sixth lunar month, Liu Yan, King of Zhongshan, passed away.) \\
\hline
\end{tabular}
\end{center}
\label{tab:example}
\end{table}
%%%%%%%%%%%%%%%%%%%%%%%%%%%%%%%%%%%%%%%%%%%%%%%%%%%%%%%%%%%%%%%%

\subsection{Translating Result}

Under most circumstances, our models can translate sentences between ancient Chinese and contemporary Chinese properly. For instance in Table~\ref{tab:example}, our models can translate ``\begin{CJK}{UTF8}{gbsn}薨\ (pass away)\end{CJK}'' into ``\begin{CJK}{UTF8}{gbsn}去世\ (pass away)\end{CJK}'' or ``\begin{CJK}{UTF8}{gbsn}逝世\ (pass away)\end{CJK}'', which are the correct forms of expression in contemporary Chinese. 
And our models can even complete some omitted characters. For instance, the family name ``\begin{CJK}{UTF8}{gbsn}刘 (Liu)\end{CJK}'' in ``\begin{CJK}{UTF8}{gbsn}中山王刘焉 (Liu Yan, King of Zhongshan)\end{CJK}'' was omitted in ancient Chinese because ``\begin{CJK}{UTF8}{gbsn}中山王 (King of Zhongshan)\end{CJK}'' was a hereditary peerage offered to ``\begin{CJK}{UTF8}{gbsn}刘 (Liu)\end{CJK}'' family. And our model completes the family name ``\begin{CJK}{UTF8}{gbsn}刘 (Liu)\end{CJK}'' when translating. For proper nouns, the seq-to-2seq baseline model can fail sometimes while the copying model can correctly copy them from the source language. For instance, the seq-to-seq baseline model translates ``\begin{CJK}{UTF8}{gbsn}焉 (Yan)\end{CJK}'' into ``\begin{CJK}{UTF8}{gbsn}刘裕 (Liu Yu, a famous figure in the history)\end{CJK}'' because ``\begin{CJK}{UTF8}{gbsn}焉 (Yan)\end{CJK}'' is relatively low-frequent words in ancient Chinese. However, the copying model learns to copy these low-frequency proper nouns from source sentences directly.

Translating dates between ancient and contemporary Chinese calendar requires background knowledge of the ancient Chinese lunar calendar, and involves non-trivial calculation that even native speakers cannot translate correctly without training.
In the example, ``\begin{CJK}{UTF8}{gbsn}辛卯\ (The Xinmao Day)\end{CJK}'' is the date presented in the ancient form, our model fails to translate it.
Our model fails to transform between the Gregorian calendar and the ancient Chinese lunar calendar and choose to generate a random date, which is expected because of the difficulty of such problems.

\section{Conclusion}
In this paper, we propose an unsupervised algorithm to construct sentence-aligned sentence pairs out of a passage-aligned corpus to build a large sentence-aligned corpus. We propose to apply the sequence to sequence model with attention and copying mechanism to automatically translate between two styles of Chinese sentences. The experimental results show that our method can yield good translating results. 

\bibliographystyle{splncs04.bst}
\bibliography{mybib.bib}
\end{document}